\documentclass[conference]{IEEEtran}
\IEEEoverridecommandlockouts

\usepackage{cite}
\usepackage{amsmath,amssymb,amsfonts}
\usepackage{algorithmic}
\usepackage{graphicx}
\usepackage{textcomp}
\usepackage{xcolor}
\usepackage{subcaption}
\usepackage{url}

\usepackage[table]{xcolor}
\usepackage{booktabs}
\usepackage{tabularx}
\usepackage{array}

\newcommand{\yescell}[1]{\cellcolor[RGB]{204,235,204}#1}
\newcommand{\nocell}[1]{\cellcolor[RGB]{242,204,204}#1}

\newcommand{\proposedyes}[1]{\cellcolor[RGB]{120,195,120}\textbf{#1}}

\newcommand{\datasetcell}[1]{\cellcolor[RGB]{232,238,244}#1}
\newcommand{\proposedname}[1]{\cellcolor[RGB]{185,215,235}\textbf{#1}}

\definecolor{RowBlue}{RGB}{198,228,246}
\definecolor{RowGreen}{RGB}{203,234,199}
\definecolor{RowOrange}{RGB}{248,222,178}
\definecolor{RowPurple}{RGB}{224,210,239}

\def\BibTeX{{\rm B\kern-.05em{\sc i\kern-.025em b}\kern-.08em
    T\kern-.1667em\lower.7ex\hbox{E}\kern-.125emX}}
\begin{document}

\title{DARP: A Calibrated Dual-Arm RGB-D-IR Dataset for Multi-View Robotic Perception}

\author{\IEEEauthorblockN{Manish Kansana*, Mohammed Yusuf Mujawar*, Sudip Mittal, Shahram Rahimi, Noorbakhsh Amiri Golilarz*}
\IEEEauthorblockA{\textit{Department of Computer Science},
\textit{The University of Alabama},
Tuscaloosa, AL, USA}
\thanks{*These authors contributed equally to this work.}
}

\maketitle

\begin{abstract}
Robotic perception from a single viewpoint is often limited by self-occlusion and incomplete surface visibility. This paper presents DARP(Dual-Arm Robotic Perception)\footnote{\url{https://dx.doi.org/10.21227/z10k-6492}}, a calibrated dual-arm RGB-D-IR dataset for object-centered robotic perception using two independently moving eye-in-hand manipulators positioned on opposite sides of a shared tabletop workspace. Each arm carries an Intel RealSense sensor that continuously records RGB, depth, and stereo infrared data while synchronized robot joint states are logged for pose recovery. Objects are placed without fixed poses or marked locations, and the acquisition procedure performs automatic localization, cross-arm confirmation, adaptive viewpoint generation, and continuous multimodal recording. DARP contains ten unique tabletop objects and preserves the original sensor recordings, robot-state logs, object-level metadata, and calibration information required to reconstruct camera trajectories in a shared metric frame. To evaluate the geometric consistency of the acquisition, we implement a deterministic multi-view fusion pipeline that converts calibrated RGB-D observations into complementary partial point clouds and measured surface meshes without using learned or generative completion methods. Evaluation on 224 held-out RGB-D keyframes comprising 1,563,466 three-dimensional query points yields a median point-to-mesh distance of 2.13~mm and an RMSE of 4.04~mm, with 96.56\% of points within 10~mm of the measured-surface mesh. DARP is intended as a reusable resource for multi-view reconstruction, collaborative robotic perception, multimodal fusion, active perception, and future learning-based reasoning over partial object observations.
\end{abstract}

\begin{IEEEkeywords}
dual-arm robotics, RGB-D dataset, eye-in-hand perception, multi-view fusion, collaborative perception, geometric calibration, point-cloud fusion, active perception
\end{IEEEkeywords}

\section{Introduction}

Robotic perception is often limited by partial visibility. A wrist-mounted camera may observe only one side of an object because of self-occlusion, object geometry, or manipulator workspace constraints. This limitation becomes important during manipulation, where information from surfaces outside the current field of view may still be useful for object understanding, grasp planning, and interaction. Multiple robotic viewpoints can reduce this limitation, but their observations must be temporally associated and expressed in a common spatial frame before they can be combined reliably.

Several datasets have supported RGB-D object recognition, multi-view perception, manipulation, and 3D reconstruction. The Washington RGB-D Object Dataset, BigBIRD, and the YCB Object and Model Set provide multi-view observations of physical objects and have been widely used for object-centric perception research \cite{lai2011rgbd,singh2014bigbird,calli2015ycb}. Robotics-oriented datasets such as GraspNet and ROBI further connect RGB-D sensing with grasping and multi-view robotic perception \cite{fang2020graspnet,yang2021robi}. Collaborative perception studies have also shown the benefit of combining observations from multiple robotic agents rather than relying on a single viewpoint \cite{reily2021collaborative,zhou2022gnn,li2022scenecompletion}. Our work focuses on a close-range object-centric setting in which two independently moving eye-in-hand manipulators observe the same physical object from complementary sides while robot motion and multimodal sensor data remain geometrically linked.

We present DARP (Dual-Arm Robotic Perception), a calibrated dual-arm robotic perception dataset collected using two Unitree Z1 Pro manipulators positioned on opposite sides of a shared tabletop workspace \cite{darp_dataset}. Each arm carries a wrist-mounted Intel RealSense camera that continuously records RGB, depth, and stereo infrared observations while both robot's joint configurations are logged with synchronized timestamps. Objects are placed without a fixed pose, marked location, or turntable. The system automatically localizes the object, confirms its position across both arms, generates object-centered viewpoints, and records complementary observations throughout the survey.

A key feature of DARP is the preservation of the geometric relationship between sensing and robot motion. ChArUco-based calibration is used to establish the required rigid transformations \cite{an2018charuco}. Camera-to-wrist hand-eye calibration relates each sensor to its corresponding manipulator, while base-to-base calibration places both robot systems in a common world frame. Recorded joint states, joint-zero corrections, and forward kinematics are then used to recover the world-frame pose of each selected camera observation. The dataset therefore retains the information needed to reconstruct metric camera trajectories and combine observations from both arms without requiring the calibration target during normal object collection.

To demonstrate the geometric consistency and practical use of DARP, we implement a deterministic dual-arm fusion pipeline. Calibrated RGB-D observations are converted to metric point clouds, transformed into the shared world frame, filtered using object support and depth consistency, and accumulated across multiple viewpoints. Repeatedly supported measurements are retained and converted into a measured surface mesh using multi-scale ball pivoting. No trained model, pretrained network, or generative method is used, and geometry that is not observed by either arm is not inferred. DARP is intentionally preserved as a task-flexible acquisition resource rather than being tied to a single reconstructed mesh or semantic benchmark. Its calibrated RGB, depth, infrared, and robot-pose information can support visible-surface reconstruction, segmentation, classification, collaborative perception, sensor-fusion studies, and next-best-view planning. The complementary partial observations may also support future learning-based shape completion approaches such as PCN and PoinTr when an appropriate training target or self-supervised objective is available \cite{yuan2018pcn,yu2021pointr}.

The main contributions of this work are:
\begin{itemize}
    \item A calibrated dual-arm perception dataset containing synchronized RGB, metric depth, stereo infrared, and robot-state observations from two independently moving eye-in-hand manipulators.
    
    \item A shared-world geometric framework combining ChArUco-based hand-eye and base-to-base calibration with recorded joint states, joint-zero corrections, and robot forward kinematics.
    
    \item An autonomous object-centered acquisition procedure that supports unconstrained object placement, cross-arm confirmation, adaptive viewpoint generation, and continuous multimodal recording.
    
    \item A deterministic measured-surface fusion pipeline with held-out RGB-D evaluation, achieving a median point-to-mesh distance of 2.13~mm over more than 1.56 million unused 3D query points.
    
    \item A task-flexible dataset design that supports both geometric processing and future learning-based applications, including segmentation, classification, collaborative perception, active perception, and partial-shape completion.
\end{itemize}

The remainder of this paper is organized as follows. Section~II reviews
related RGB-D datasets, partial-view reconstruction, collaborative
robotic perception, and calibration-based geometric processing.
Section~III presents the dual-arm dataset generation and geometric
processing framework, including the robotic platform, calibration,
autonomous acquisition, dataset organization, and deterministic
fusion procedure. Section~IV reports dataset characteristics and
quantitative and qualitative results. Section~V discusses supported
applications and compares DARP with existing resources.
Section~VI summarizes limitations and practical considerations, and
Section~VII concludes the paper.

\section{Related Work}

\subsection{RGB-D and Multi-View Object Datasets}

RGB-D datasets have been widely used for object recognition, manipulation, and 3D perception. The Washington RGB-D Object Dataset provides multi-view RGB-D observations of physical objects and established an early benchmark for object-centric RGB-D recognition \cite{lai2011rgbd}. BigBIRD further increased viewpoint coverage for real object instances through dense RGB-D acquisition \cite{singh2014bigbird}, while the YCB Object and Model Set introduced a standardized collection of physical objects and geometric models for robotic manipulation research \cite{calli2015ycb}. Large-scale resources such as ShapeNet and ScanNet address complementary settings through digital 3D models and reconstructed indoor RGB-D scenes, respectively \cite{chang2015shapenet,dai2017scannet}.

More robotics-oriented datasets directly connect visual sensing with manipulation. GraspNet-1Billion provides RGB-D observations and large-scale grasp annotations for general object grasping \cite{fang2020graspnet}. ROBI focuses on multi-view sensing of reflective objects in robotic bin-picking scenarios and highlights the difficulty of acquiring reliable geometry from challenging surfaces \cite{yang2021robi}. These datasets provide valuable resources for robotic perception, but their acquisition goals differ from continuously recording complementary observations from two independently moving eye-in-hand manipulators together with synchronized robot states.

\subsection{Partial-View Reconstruction and Shape Completion}

Partial observations naturally arise when objects are viewed from limited directions or contain self-occluded surfaces. Learning-based methods have therefore explored how incomplete point clouds can be mapped to more complete shape representations. PCN introduced an encoder--decoder architecture for point-cloud completion \cite{yuan2018pcn}, while TopNet proposed a hierarchical structural decoder \cite{tchapmi2019topnet}. More recent methods such as PoinTr and SnowflakeNet use Transformer-based or progressive point-generation mechanisms to model missing geometry from partial observations \cite{yu2021pointr,xiang2021snowflakenet}. ProxyFormer and AnchorFormer further develop Transformer-based completion using missing-part reasoning and discriminative geometric structures \cite{li2023proxyformer,chen2023anchorformer}.

Generative approaches have also been investigated for point-cloud completion, including diffusion-based methods such as PCDreamer and SuperPC \cite{wei2025pcdreamer,du2025superpc}. Recent work has additionally emphasized that performance on synthetic or controlled completion benchmarks does not necessarily translate directly to real sensor observations \cite{pathak2025realworldcompletion}. These methods motivate future learning applications of the proposed dataset; however, the reconstruction presented in this paper is strictly geometric and does not infer unseen object surfaces.

\subsection{Collaborative Robotic Perception}

Collaborative perception uses observations from multiple agents to reduce the limitations of an individual viewpoint. Reily and Zhang studied joint view and feature selection for collaborative multi-robot perception \cite{reily2021collaborative}, while Zhou \textit{et al.} explored graph neural networks for combining perceptual information across robots \cite{zhou2022gnn}. Multi-Robot Scene Completion further considered collaborative perception as a task-agnostic representation problem in which information from multiple robots contributes to a more complete scene description \cite{li2022scenecompletion}. Recent datasets such as CU-Multi also provide synchronized observations for multi-robot collaborative perception research \cite{albin2026cumulti}.

The setting considered in this work is more object-centered and close-range. Two fixed-base manipulators independently move eye-in-hand cameras around the same tabletop object, producing complementary partial observations while robot kinematics provide the pose of each sensor measurement. This makes the dataset suitable for studying how embodied viewpoints can be combined at the level of object geometry as well as for later feature-level collaborative perception.

\subsection{Calibration and Geometric Fusion}

Reliable fusion of observations from moving robot-mounted cameras requires accurate spatial calibration. The hand-eye calibration problem estimates the rigid transformation between the robot end-effector and the attached camera, with classical solutions proposed by Tsai and Lenz and later by Daniilidis \cite{tsai1989handeye,daniilidis1999dualquaternions}. ChArUco patterns combine coded ArUco markers with chessboard corners and provide feature correspondences that can be used for intrinsic and extrinsic calibration \cite{an2018charuco}. In the proposed system, ChArUco-based calibration is used to establish the camera-to-wrist and inter-base relationships required to place observations from both arms in a common metric frame.

Multi-view RGB-D reconstruction methods such as KinectFusion and BundleFusion demonstrate how repeated depth observations can be integrated into coherent geometry when reliable camera poses are available \cite{newcombe2011kinectfusion,dai2017bundlefusion}. Our geometric pipeline follows the same general principle of transforming repeated measurements into a shared frame, but the camera poses are derived primarily from robot calibration, synchronized joint states, and forward kinematics rather than estimated through visual tracking.

Overall, existing work provides strong resources for RGB-D perception, robotic manipulation, point-cloud completion, collaborative perception, and multi-view reconstruction. The proposed dataset connects these areas through a dual-arm eye-in-hand acquisition setting that preserves multimodal observations, robot motion, and calibration information in a shared metric coordinate system.

\section{Dual-Arm Dataset Generation Framework}

The dataset is created using a calibrated dual-arm robotic acquisition system designed to collect complementary RGB-D-IR observations of tabletop objects. Two fixed-base manipulators observe the same object from opposite sides of a shared workspace, while the corresponding robot joint states and calibration parameters are retained so that each sensor observation can later be placed in a common metric coordinate frame. The complete framework consists of four main components namely, the robotic sensing platform, spatial calibration and coordinate alignment, autonomous object-centered acquisition, and preservation of the resulting multimodal data and robot metadata.

\subsection{Robotic Platform and Sensors}

\begin{figure}[t]
    \centering
    \includegraphics[width=\columnwidth]{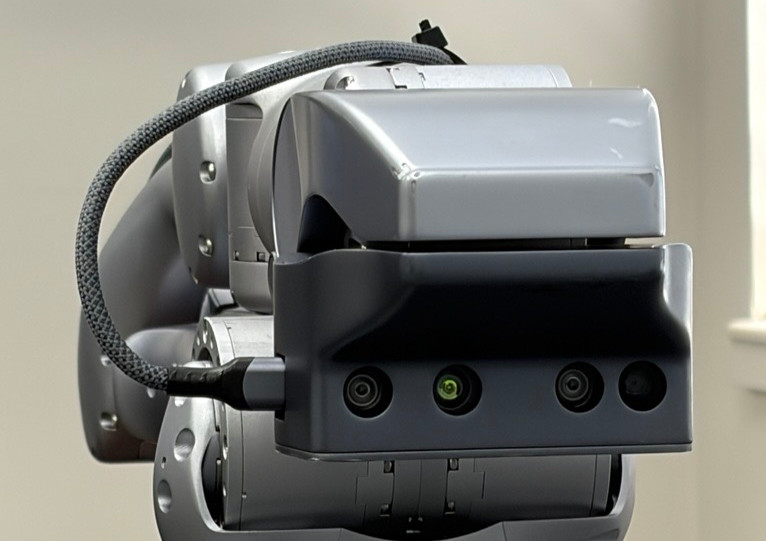}
    \caption{Intel RealSense RGB-D-IR sensor mounted near the
gripper flange of each Unitree Z1 Pro manipulator.}
    \label{fig:sensors}
\end{figure}

\begin{figure}[t]
    \centering
    \includegraphics[width=\columnwidth]{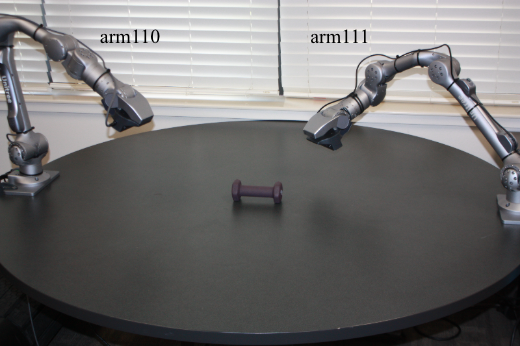}
    \caption{Dual-arm acquisition platform. Two Unitree Z1 Pro manipulators are mounted on opposite sides of the tabletop workspace.}
    \label{fig:platform}
\end{figure}

The acquisition platform consists of two 6-DoF Unitree Z1 Pro
manipulators, denoted as arm110 (one on the left) and arm111 (one on the right), mounted on opposite sides
of a shared tabletop workspace, as shown in
Fig.~\ref{fig:platform}. The two robot bases remain fixed throughout
data collection and are oriented approximately toward one another,
creating an overlapping manipulation region in which both arms can
observe the same physical object from different directions.

Each manipulator carries an Intel RealSense RGB-D camera rigidly
attached near the gripper flange in an eye-in-hand configuration, as
shown in Fig.~\ref{fig:sensors}. Camera serial 348522076490 is
associated with arm110, while camera serial 405622074504 is associated
with arm111. This mapping is fixed because every camera observation must later be transformed through the kinematic chain of its corresponding arm. The cameras continuously record color, depth, and stereo infrared information at $640\times480$ resolution and 30~Hz. Color is stored in BGR8 format, depth in Z16 format, and the left and right infrared channels in Y8 format. Accelerometer and gyroscope streams are also retained in the native recordings, although they are not used by the geometric reconstruction presented in this work. Camera intrinsics, inter-stream calibration, and depth scale remain embedded in the RealSense recordings and are read from each device or bag file during processing rather than being hard-coded.

The two manipulators are separated by approximately 1.3986~m. Their relative orientation is close to antiparallel, with an approximately $178.4^{\circ}$ rotation about the world $z$ axis. This arrangement allows the two eye-in-hand cameras to observe complementary regions of the object surface while remaining within their respective reachable workspaces.

\subsection{Calibration and Coordinate Alignment}

\begin{figure}[t]
    \centering
    \includegraphics[width=\columnwidth]{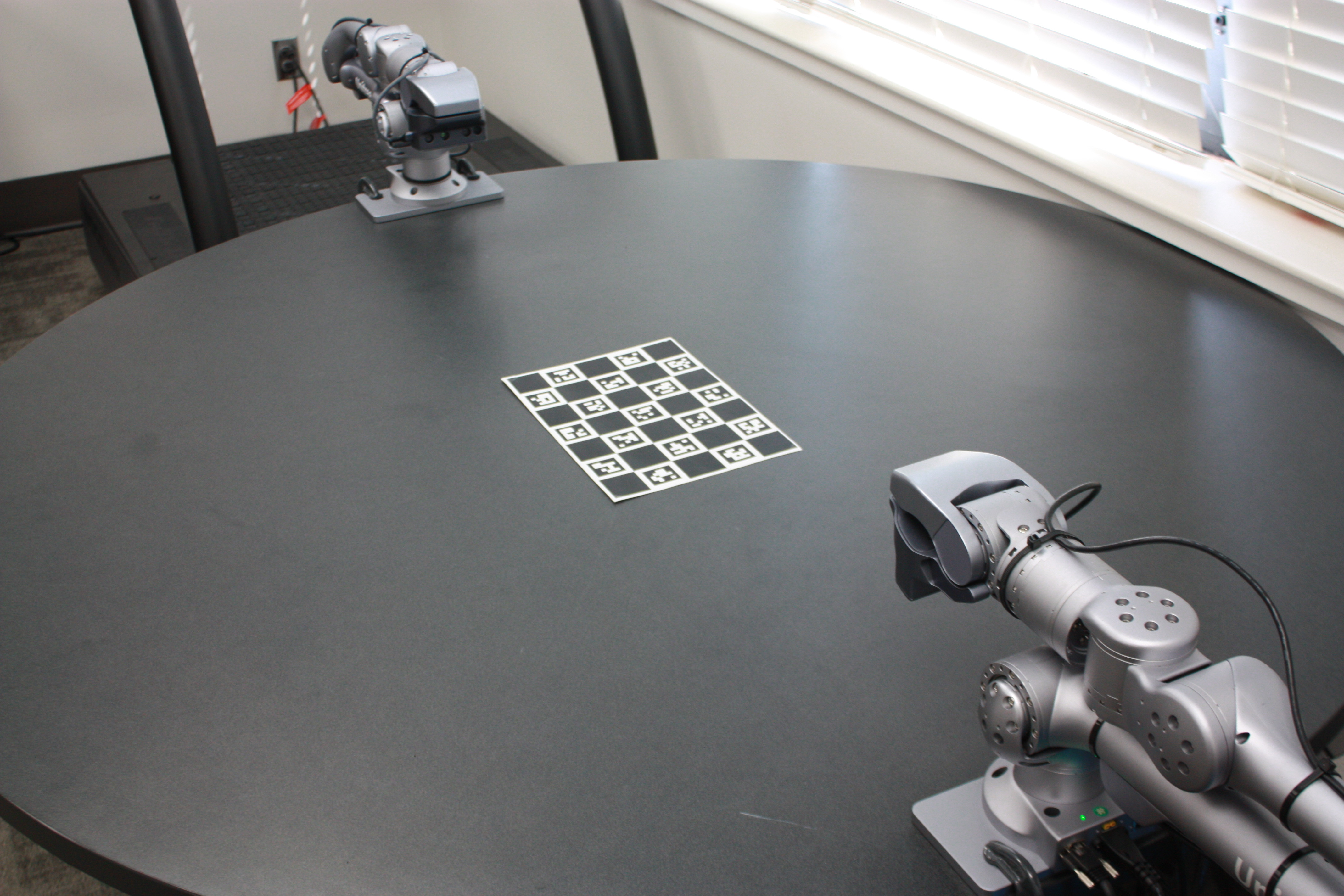}
    \caption{ChArUco-based calibration setup used to establish the
    eye-in-hand camera transformations and the geometric relationship
    between the two fixed robot bases.}
    \label{fig:charuco_calibration}
\end{figure}

Because both cameras move with their corresponding manipulators, multi-view observations can only be combined reliably when the relationship between the camera, robot wrist, robot base, and shared world frame is known. We define the world frame as $\mathcal{F}_{W}$, the base frame of arm $i$ as $\mathcal{F}_{B_i}$, the wrist or gripper frame as $\mathcal{F}_{G_i}$, and the camera optical frame as $\mathcal{F}_{C_i}$. A ChArUco board is used as the common calibration target, as shown in
Fig.~\ref{fig:charuco_calibration}. ChArUco patterns combine coded
markers with accurately localized chessboard corners and provide
feature correspondences for geometric calibration
\cite{an2018charuco}. Multiple observations of the calibration target are collected from different robot configurations to estimate the rigid camera-to-wrist transformation for each eye-in-hand sensor. The same calibration framework is used to establish the spatial relationship between the two fixed robot bases.

For each arm, two static transforms are retained, the wrist-to-camera transform ${}^{G_i}\mathbf{T}_{C_i}$ and the base-to-world transform ${}^{W}\mathbf{T}_{B_i}$. Robot joint-zero corrections are also applied before forward kinematics to compensate for static joint bias. The world-frame pose of camera $i$ at time $t$ is therefore recovered as

\begin{equation}
{}^{W}\mathbf{T}_{C_i}(t)
=
{}^{W}\mathbf{T}_{B_i}
\;
{}^{B_i}\mathbf{T}_{G_i}
\left(\mathbf{q}_i(t)+\Delta\mathbf{q}_i\right)
\;
{}^{G_i}\mathbf{T}_{C_i},
\label{eq:world_camera_pose}
\end{equation}

where $\mathbf{q}_i(t)$ denotes the interpolated joint configuration and $\Delta\mathbf{q}_i$ contains the static joint-zero correction for arm $i$.

The robot-state logger operates at approximately 33~Hz, while the cameras record at 30~Hz. For each camera timestamp, the surrounding robot states are linearly interpolated to obtain the corresponding joint configuration. The robot log begins before camera acquisition and ends after recording is complete so that the usable camera interval remains inside the valid robot-pose interval and pose extrapolation is not required.

The final hand-eye calibration reports held-out geometric errors of approximately 3.58~mm and 3.93~mm for the two arms. These values characterize the internal calibration consistency and are later complemented by cross-arm and held-out RGB-D evaluation. The calibration is treated as epoch-specific: if a camera is remounted, a robot base is moved, the sensor bracket changes, or joint-zero behavior changes, the corresponding calibration should be repeated.

\subsection{Autonomous Data Collection}

Each object survey begins when the operator places a single object somewhere within the shared reachable workspace and starts the acquisition procedure. The object position and orientation are not manually measured, no marked placement location is required, and no turntable is used. The system therefore estimates the object location during each survey before generating the viewing trajectories.

\begin{figure}[t]
    \centering
    \includegraphics[width=1\columnwidth]
    {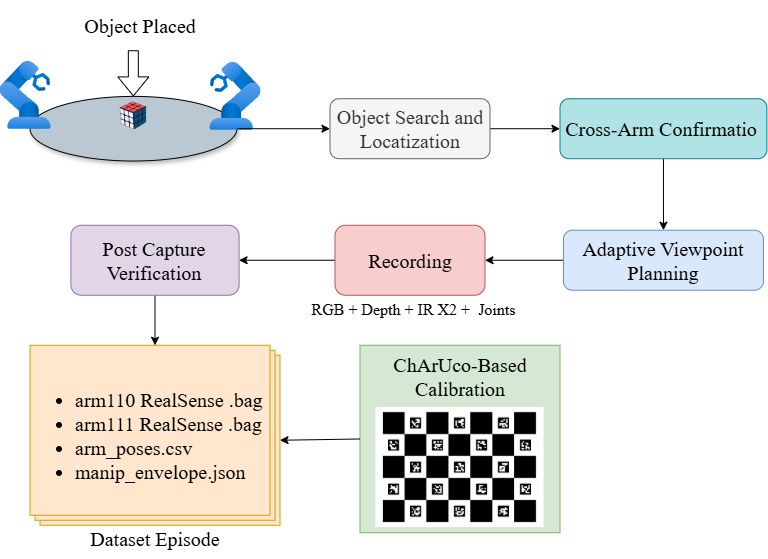}
    \caption{Overview of the dual-arm data collection pipeline. }
    \label{fig:collection_pipeline}
\end{figure}

The complete autonomous acquisition sequence is summarized in
Fig.~\ref{fig:collection_pipeline}. Following object placement, the
system performs object localization, cross-arm confirmation, adaptive
viewpoint planning, synchronized multimodal recording, and
post-capture verification before the episode is accepted into the
dataset.

Object localization begins with a structured pan-based search using the eye-in-hand camera. The searching manipulator directs the camera toward candidate regions of the tabletop until a valid depth-based object estimate is obtained. The estimated object center and spatial extent are treated as an initial hypothesis rather than being used immediately for trajectory generation. The object location is then independently checked from both manipulators. During this cross-arm confirmation stage, one arm observes the object while the other remains parked. Estimates that fall outside the valid workspace or exhibit unreliable depth support are rejected. When both arms produce consistent estimates, their object-center estimates are combined, while the larger estimated radius and hten are retained to form a conservative object envelope. Conflicting detections are not blindly averaged, the system instead falls back to a valid reference estimate or terminates the automated survey.

\begin{figure}[t]
    \centering
    \includegraphics[width=\columnwidth]
    {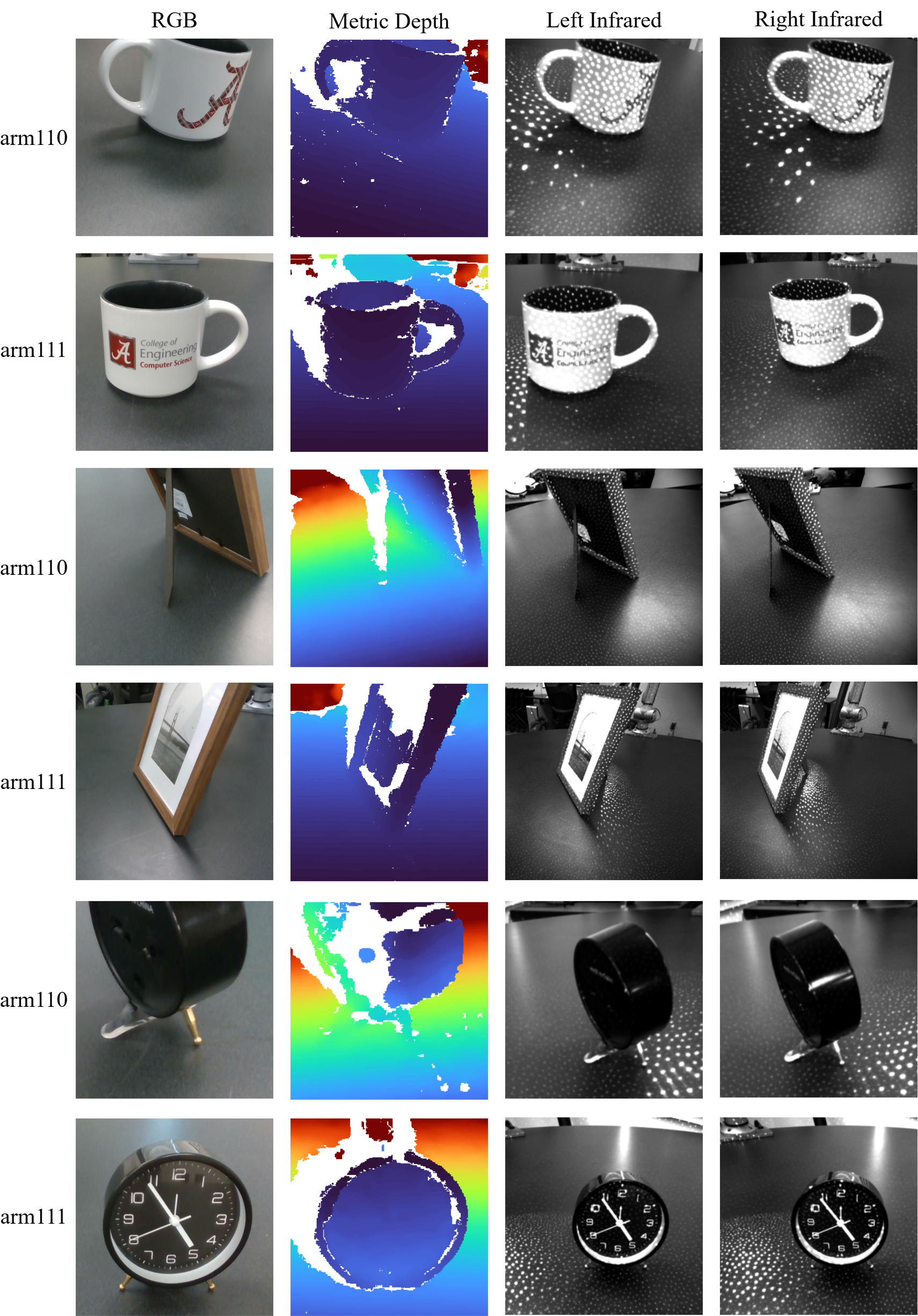}
    \caption{Representative multimodal observations of several dataset objects recorded by arm110
and arm111. Each observation contains RGB, metric depth, left
infrared, and right infrared measurements from the corresponding
eye-in-hand sensor.}
\label{fig:fullview}
\end{figure}

Using the confirmed object center and extent, an independent viewing trajectory is generated for each arm. The orbit adapts to object size, with larger objects receiving broader or denser angular coverage. Candidate camera poses are oriented toward the estimated object center and checked for inverse-kinematics feasibility, workspace limits, table clearance, excessive joint motion, and interference with the parked manipulator. Invalid candidates are removed, and the trajectory can be regenerated if too few feasible views remain. Only one manipulator moves at a time, while the other stays in a parked configuration. Both cameras record continuously during the survey, producing the
representative RGB, metric depth, and stereo infrared observations
shown in Fig.~\ref{fig:fullview}. Robot motion is executed using non-blocking commands so that joint-state logging continues during movement, and a short settling interval is applied after each accepted viewpoint.

This collection strategy is designed to produce complementary partial observations rather than identical views from both arms. Each camera therefore records a sequence of changing eye-in-hand viewpoints from its accessible side of the object, providing the geometric diversity later used for cross-arm fusion and evaluation.

\subsection{Dataset Organization}

\begin{figure*}[t!]
    \centering
    \begin{subfigure}[t]{0.19\textwidth}
        \includegraphics[width=\linewidth]{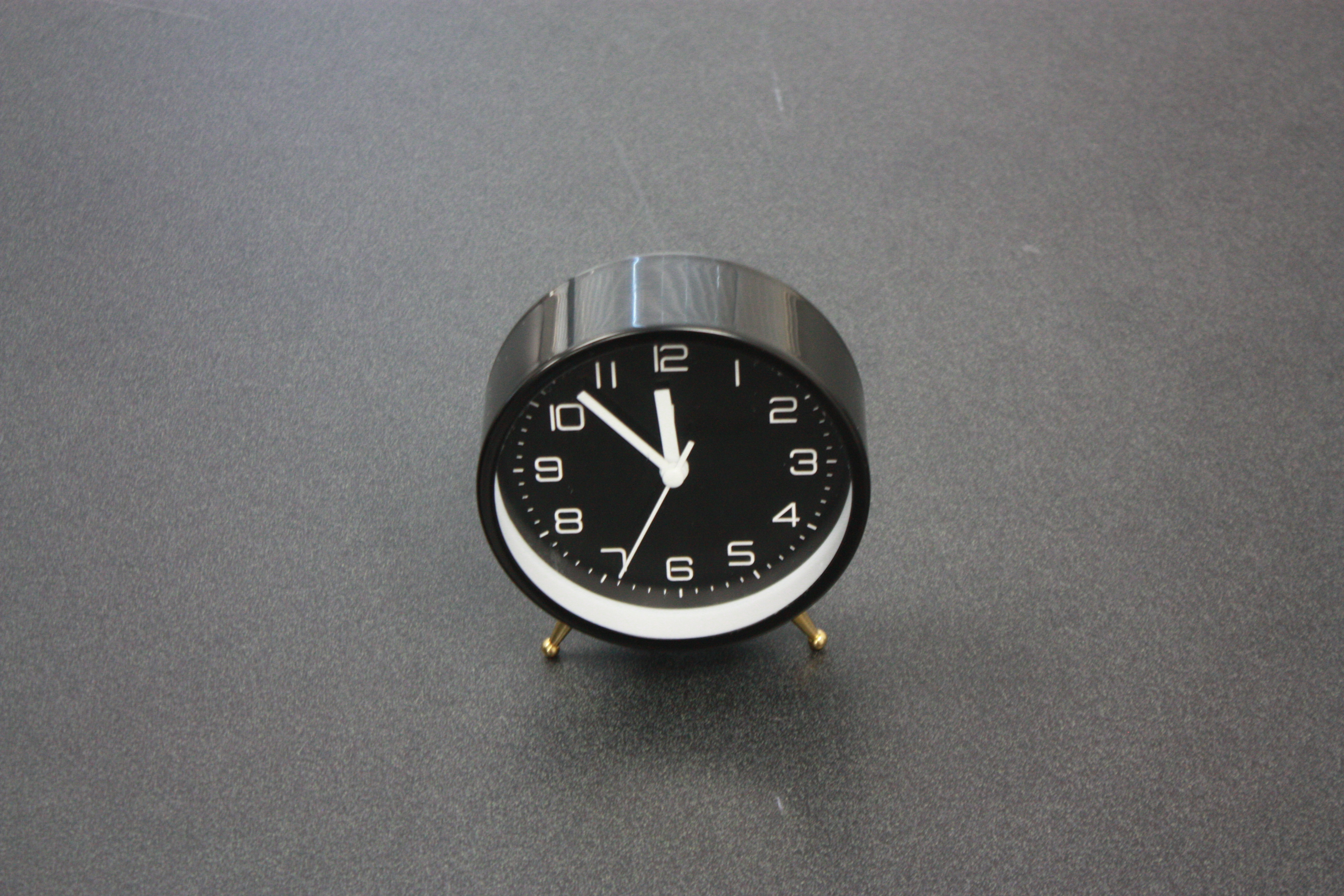}
        \caption{Clock}
    \end{subfigure}\hfill
    \begin{subfigure}[t]{0.19\textwidth}
        \includegraphics[width=\linewidth]{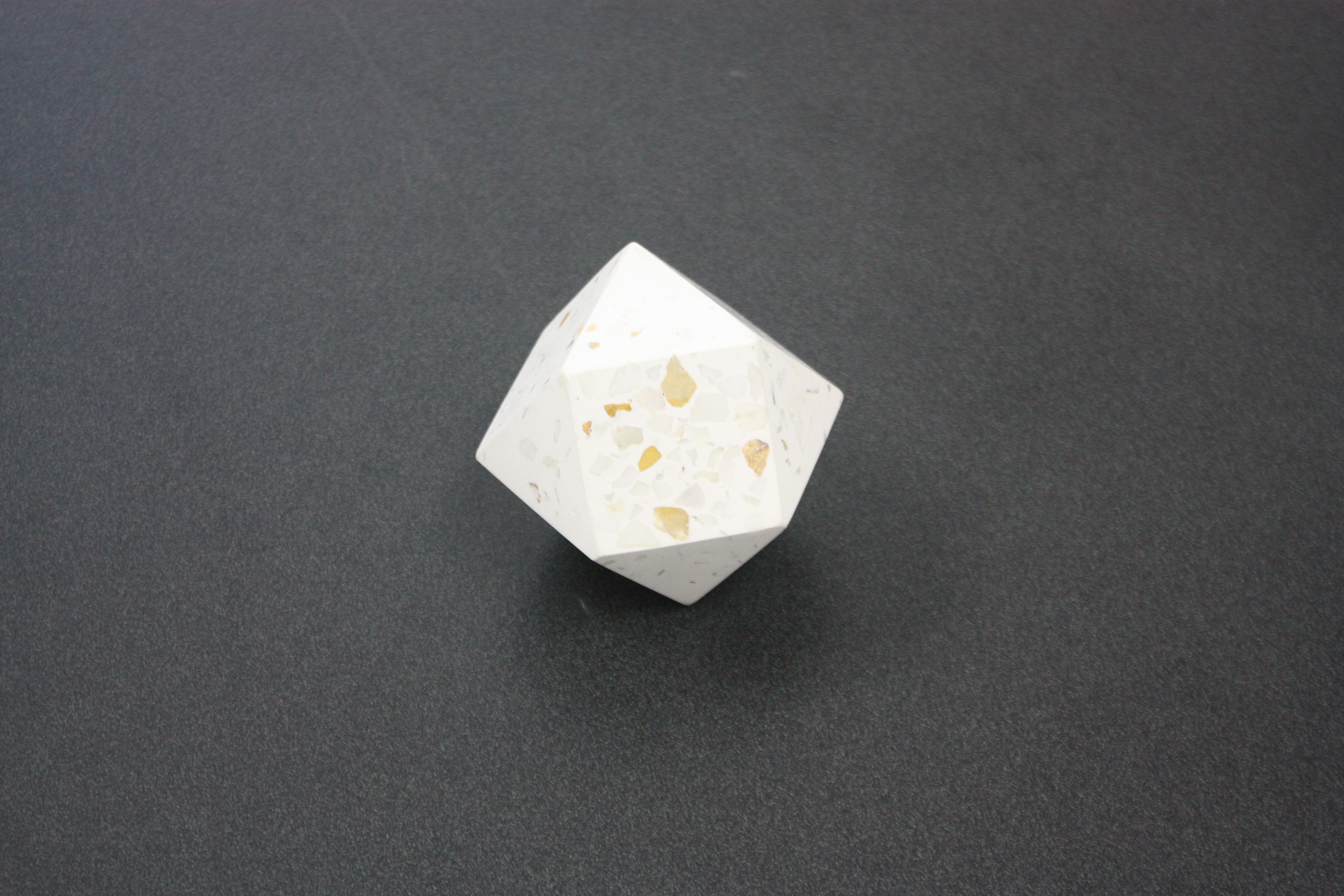}
        \caption{Weight}
    \end{subfigure}\hfill
    \begin{subfigure}[t]{0.19\textwidth}
        \includegraphics[width=\linewidth]{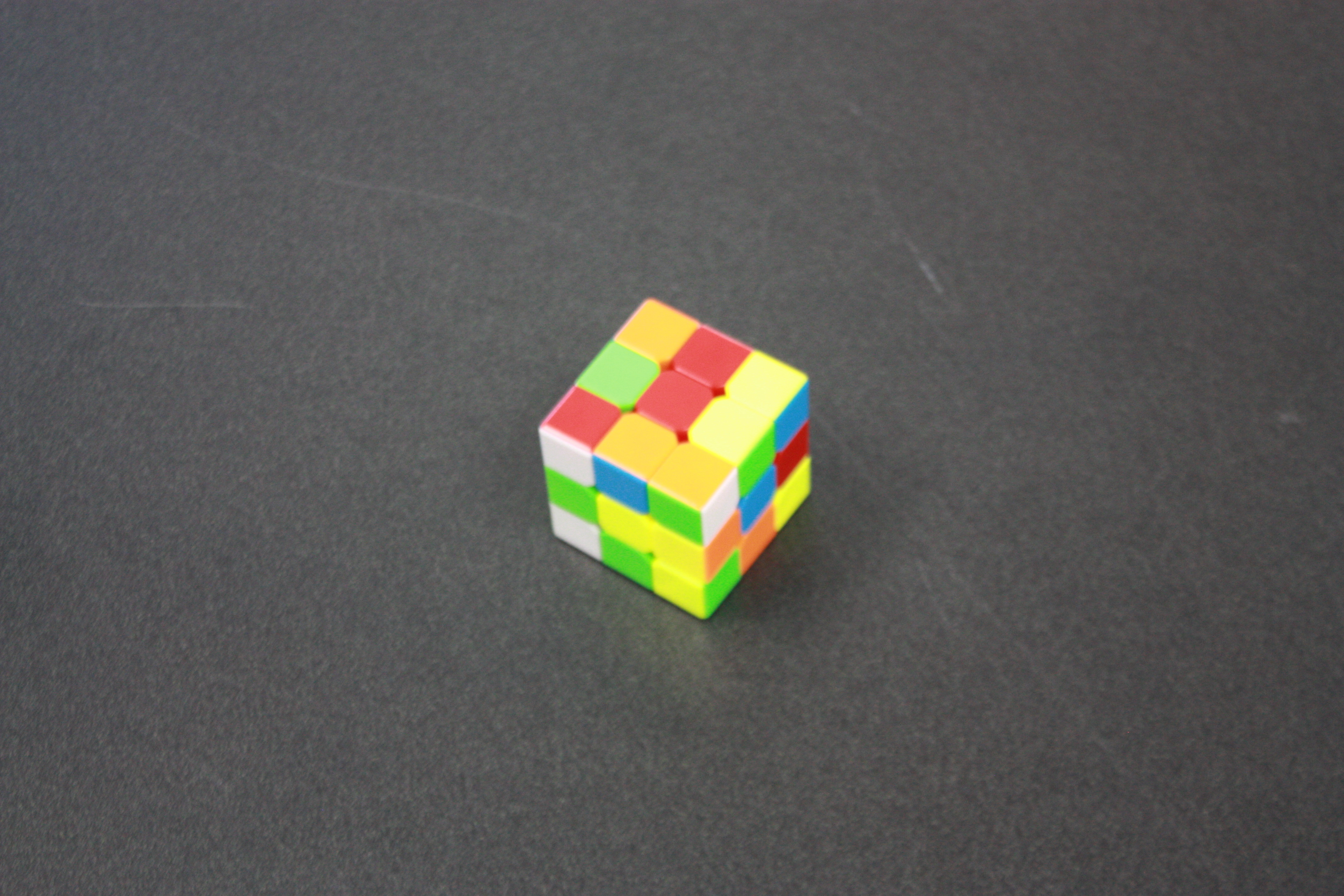}
        \caption{Cube}
    \end{subfigure}\hfill
    \begin{subfigure}[t]{0.19\textwidth}
        \includegraphics[width=\linewidth]{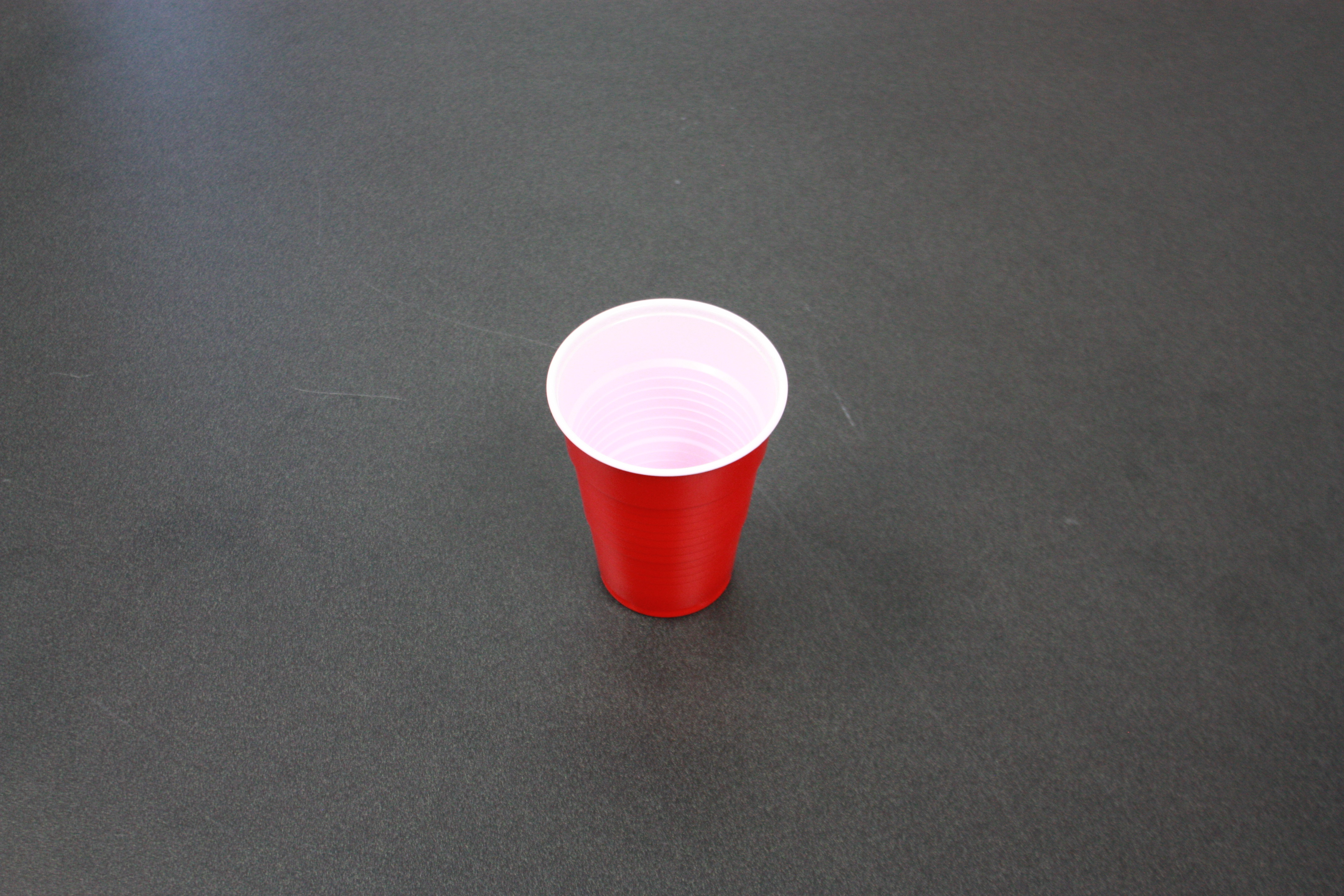}
        \caption{Cup}
    \end{subfigure}\hfill
    \begin{subfigure}[t]{0.19\textwidth}
        \includegraphics[width=\linewidth]{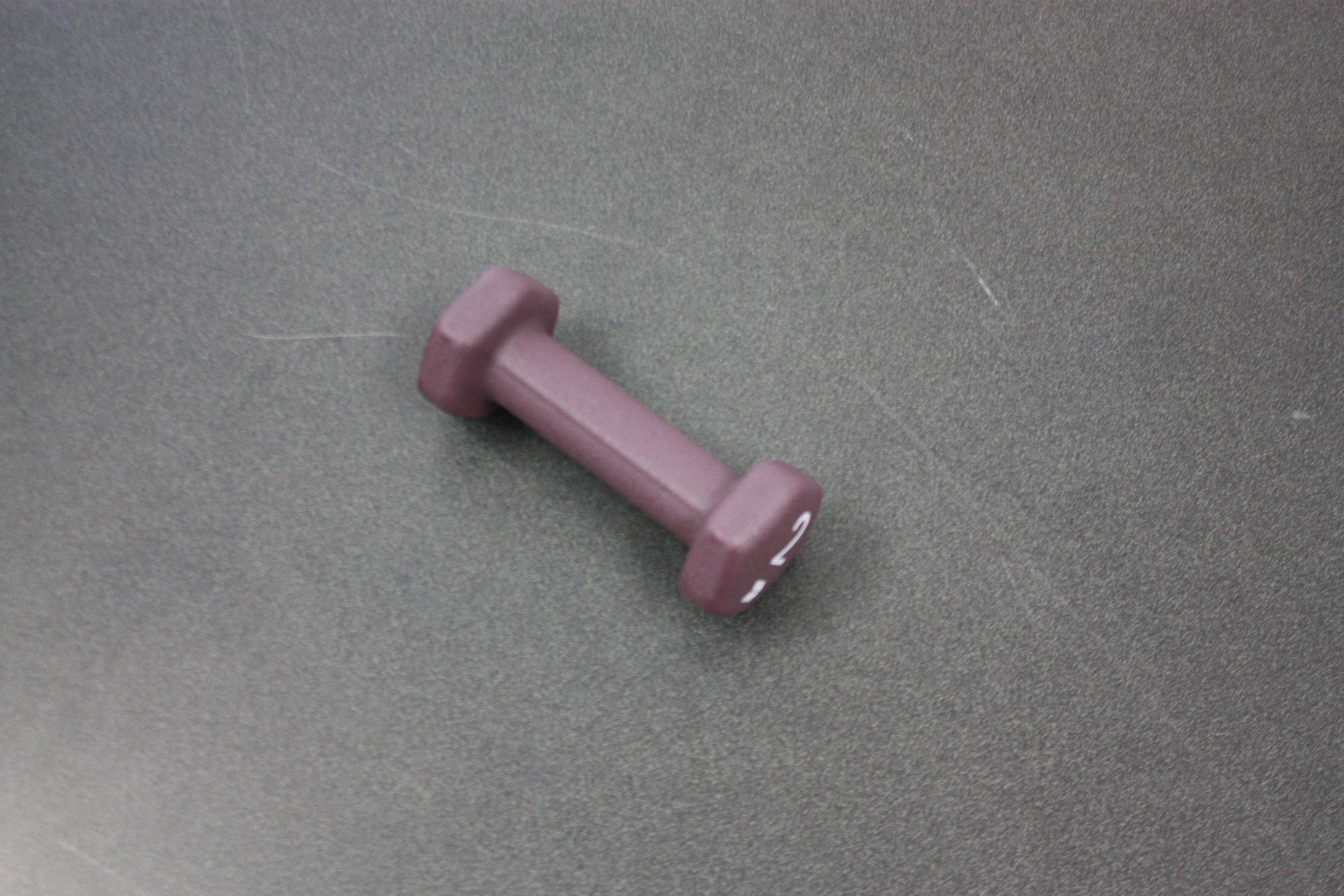}
        \caption{Dumbbell}
    \end{subfigure}

    \vspace{0.3cm}

    \begin{subfigure}[t]{0.19\textwidth}
        \includegraphics[width=\linewidth]{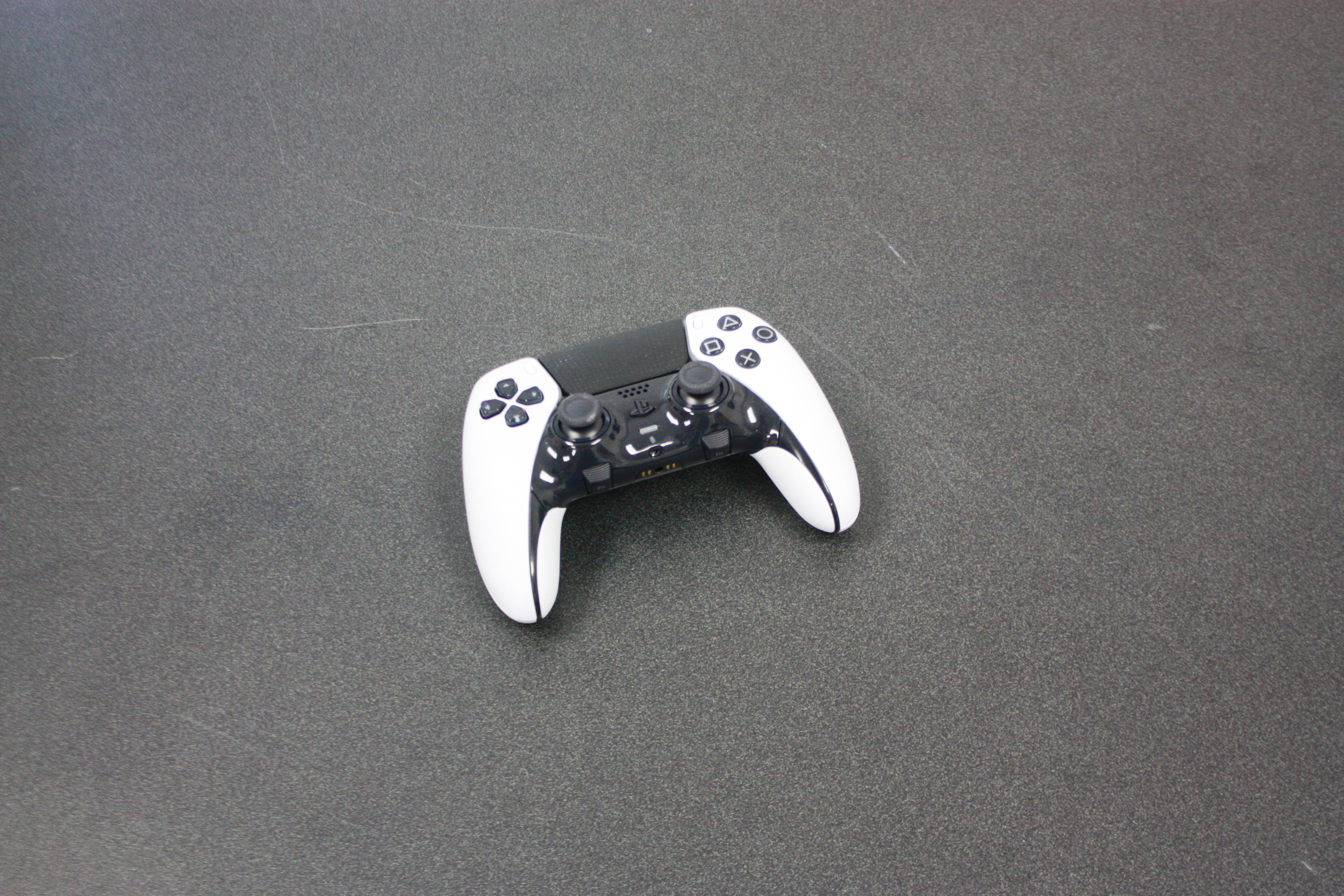}
        \caption{Game}
    \end{subfigure}\hfill
    \begin{subfigure}[t]{0.19\textwidth}
        \includegraphics[width=\linewidth]{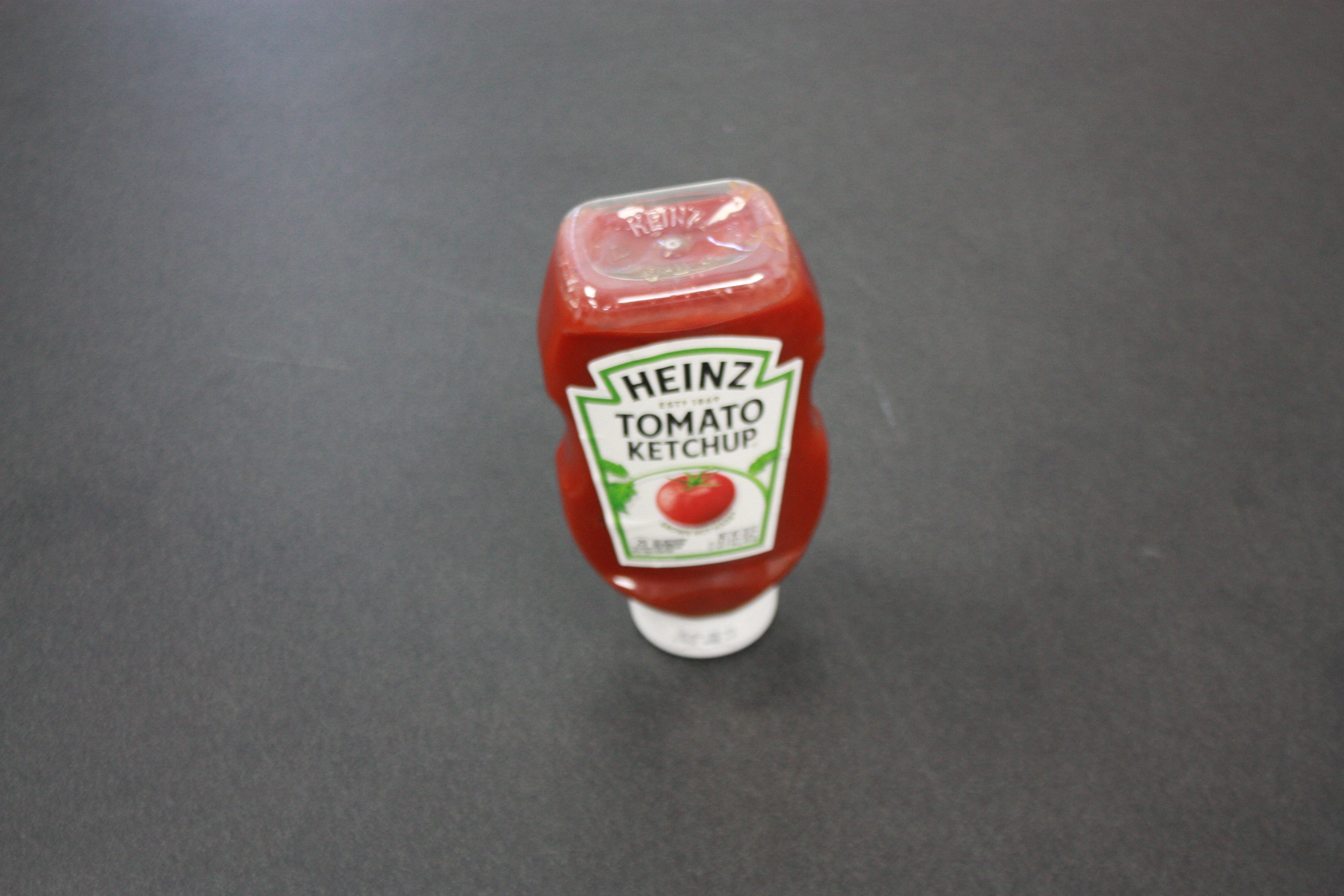}
        \caption{Ketchup}
    \end{subfigure}\hfill
    \begin{subfigure}[t]{0.19\textwidth}
        \includegraphics[width=\linewidth]{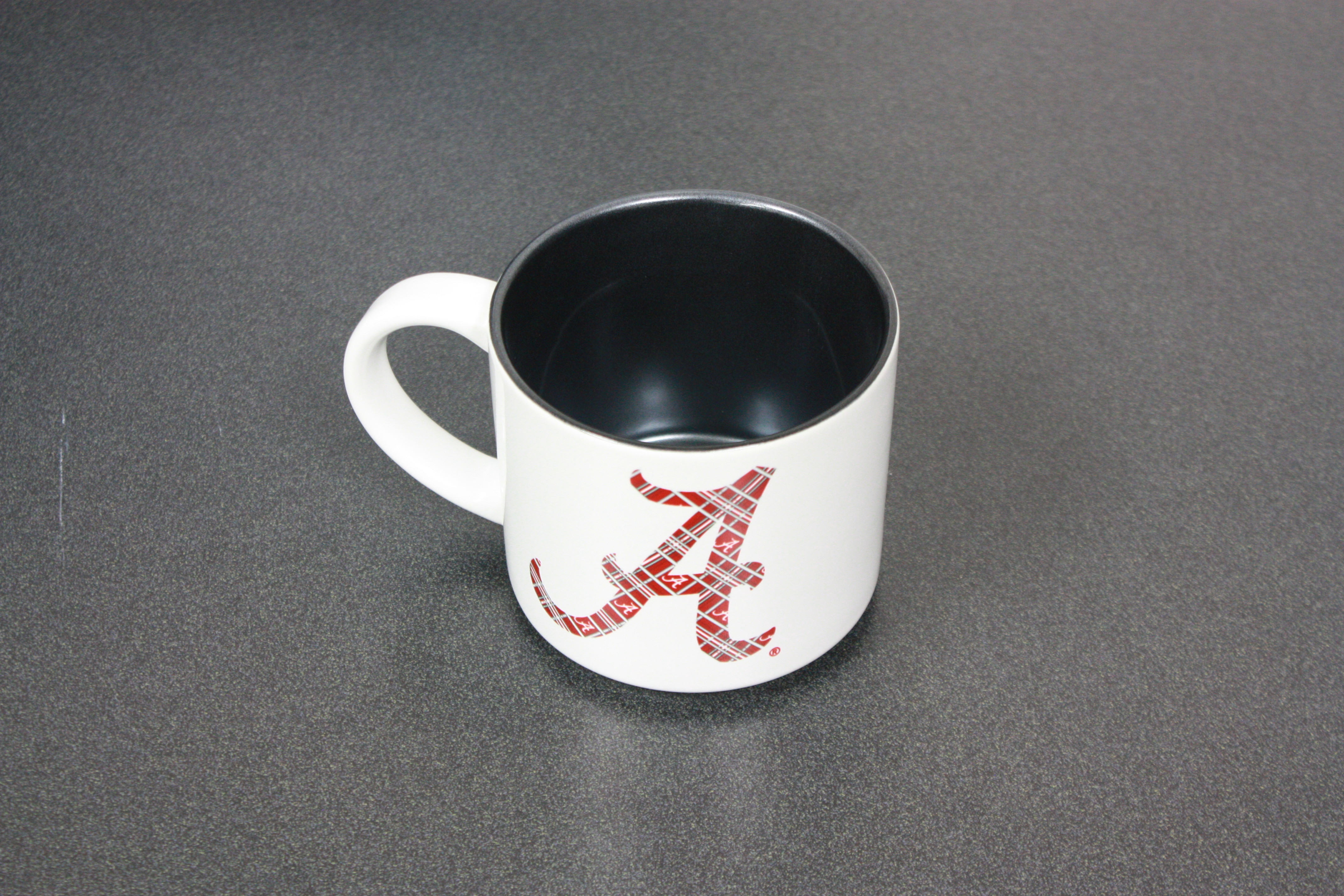}
        \caption{Mug}
    \end{subfigure}\hfill
    \begin{subfigure}[t]{0.19\textwidth}
        \includegraphics[width=\linewidth]{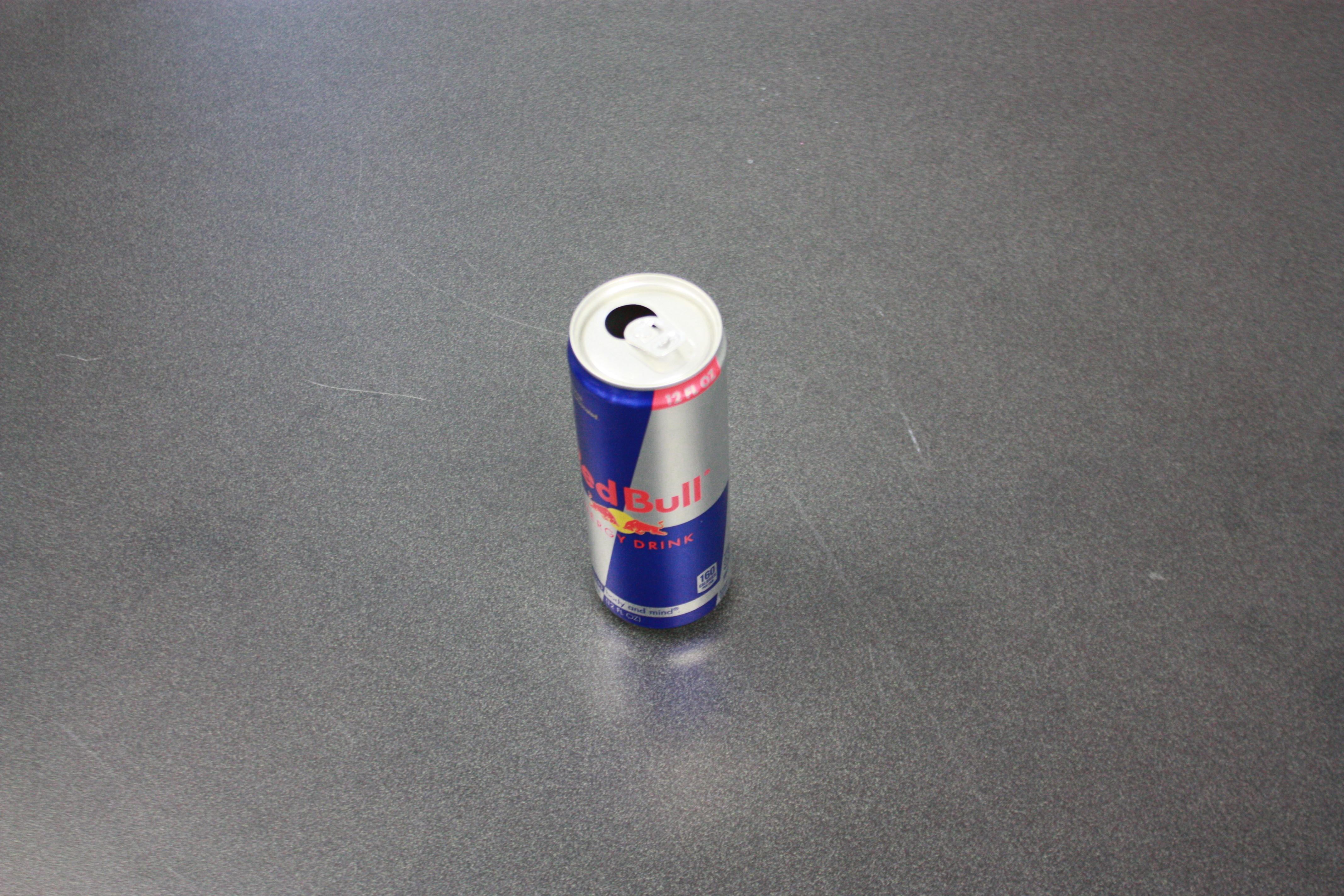}
        \caption{Red Bull}
    \end{subfigure}\hfill
    \begin{subfigure}[t]{0.19\textwidth}
        \includegraphics[width=\linewidth]{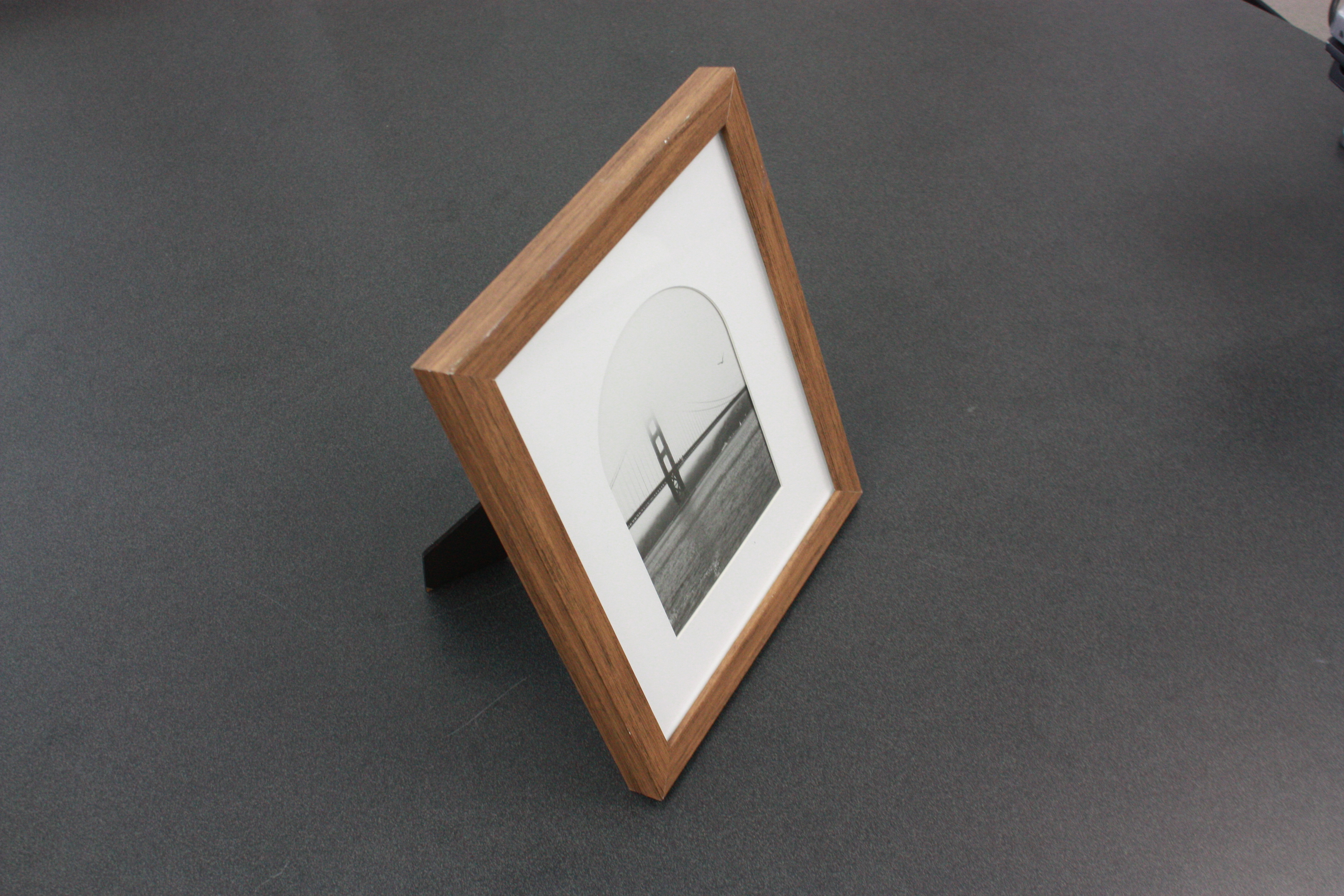}
        \caption{Frame}
    \end{subfigure}
    
    \caption{Pictures of objects used in the experiment.}
    \label{fig:objects}
\end{figure*}

Fig.~\ref{fig:objects} shows the physical tabletop objects included
in DARP. The released dataset contains ten unique physical objects. Each object survey is preserved in acquisition form and contains the two native RealSense recordings, the synchronized robot-state log, and the object-centered acquisition metadata:

\begin{verbatim}
<object>/
    cam348522076490.bag
    cam405622074504.bag
    arm_poses.csv
    manip_envelope.json
\end{verbatim}

The two \texttt{.bag} files are continuous multi-view recordings, not single images or isolated viewpoints. Each contains the RGB, depth, left infrared, and right infrared streams from one eye-in-hand camera together with the stream profiles, intrinsics, and depth scale required to decode the measurements. The \texttt{arm\_poses.csv} file stores timestamped joint states for both manipulators, including the arm identifier and six recorded joint angles. The \texttt{manip\_envelope.json} file stores the estimated object center, radius, height, nominal camera standoff, and robot configuration information used during collection.

For efficient downstream processing, selected observations can also be exported as compressed keyframes containing aligned RGB, raw depth, stereo infrared, a conservative geometric object-support mask, camera intrinsics, depth scale, and the corresponding $4\times4$ world-from-camera pose. The raw acquisition files remain the primary dataset so that alternative frame-selection, calibration, segmentation, fusion, or learning pipelines can be applied without repeating the physical experiment.

The dataset is intentionally not tied to one fixed complete object mesh or one semantic task. Instead, it preserves the multimodal sensor observations, robot motion, and calibration information required to regenerate camera trajectories, partial point clouds, fused geometric representations, or learning inputs.

\subsection{Geometric Fusion and Surface Reconstruction}

The geometric pipeline is used to demonstrate that observations from
the two eye-in-hand sensors can be combined consistently in the shared
world frame. For each selected RGB-D frame, raw depth is converted to
metric depth using the device-specific scale and back-projected using
the recorded camera intrinsics. The resulting camera-frame points are
transformed into the world frame using
Eq.~(\ref{eq:world_camera_pose}).

A conservative object envelope derived during acquisition removes
table and background measurements. View-diverse observations from
both arms are then accumulated in the shared frame. The final
production fusion uses the calibrated robot-camera poses directly,
local depth discontinuities and weakly supported voxels are rejected,
followed by statistical outlier removal.

A measured surface mesh is generated from the fused object cloud using
multi-scale ball pivoting. The procedure does not predict unobserved
geometry and therefore preserves missing or occluded regions rather
than artificially closing them. A subset of RGB-D keyframes is excluded
from fusion and retained for the held-out point-to-mesh evaluation
reported in the following section.

\section{Dataset Characteristics and Results}

This section summarizes the collected data and evaluates whether observations acquired from the two independently moving eye-in-hand cameras remain geometrically consistent after calibration and world-frame transformation. The analysis focuses on the properties that are most relevant to reuse of the dataset mainly, complementary viewpoint coverage, cross-arm alignment, and agreement between reconstructed surfaces and RGB-D observations that were not used during fusion.

\begin{figure*}[t]
    \centering
    \includegraphics[width=0.96\textwidth]
    {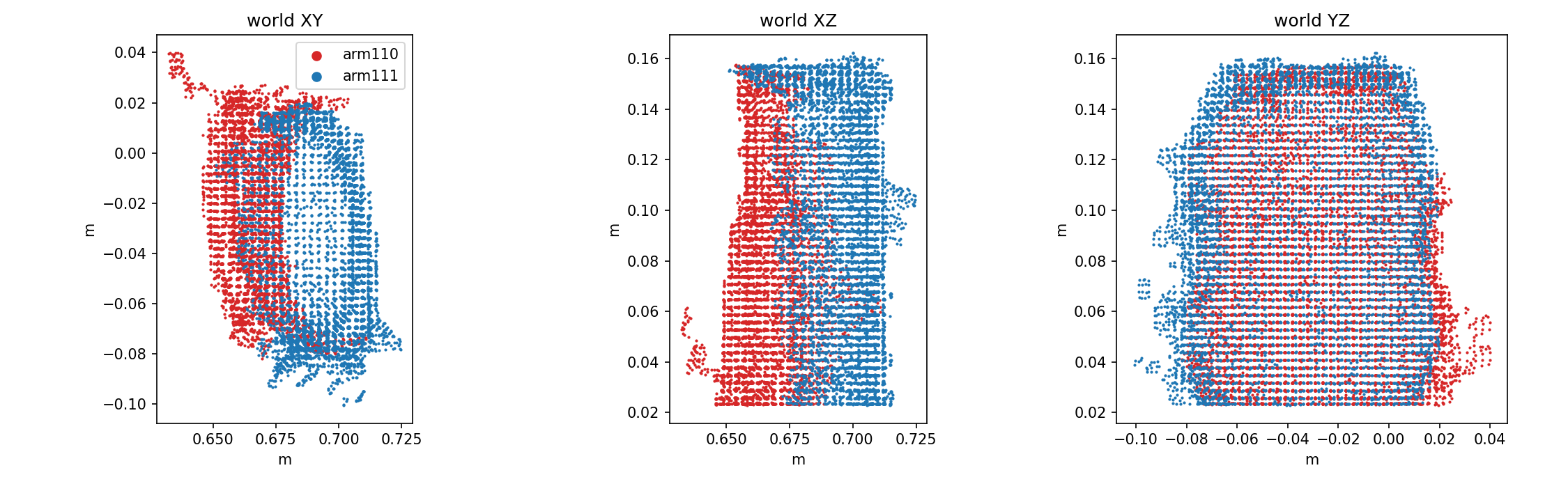}
    \caption{Cross-arm alignment of the arm110 and arm111 partial point
    clouds in the shared world frame for Ketchup object. The figure shows orthographic
    projections in the world $XY$, $XZ$, and $YZ$ planes. Red points
    correspond to arm110 and blue points correspond to arm111.}
    \label{fig:cross_arm_alignment}
\end{figure*}

\subsection{Dataset Characteristics}

The released dataset contains eight unique physical tabletop objects.
Each object is observed from both sides of the workspace using the two
eye-in-hand RealSense cameras described in Section~III. For every
object survey, the two cameras record continuous RGB, depth, and stereo
infrared streams while the joint configurations of both manipulators
are logged simultaneously. The resulting data therefore contain
multiple changing viewpoints from each arm rather than a single image
pair.

Table~\ref{tab:release_composition} summarizes the physical object
envelopes and recording statistics of the released acquisitions. Across the retained recordings, the dataset contains 61,012 robot
joint-state rows from the two manipulators and approximately
18.4~min of synchronized dual-arm acquisition. The native recordings
contain more than 63,000 depth frames from the two sensors and occupy
approximately 137.94~GB. These statistics reflect the continuous
multi-view nature of DARP, in which each episode preserves a temporal
sequence of changing eye-in-hand observations rather than a small set
of isolated viewpoints.

\begin{table*}[t]
\centering
\caption{Release composition and recording statistics for DARP.}
\label{tab:release_composition}

\definecolor{StatRow}{RGB}{238,245,250}
\rowcolors{2}{StatRow}{white}

\setlength{\tabcolsep}{4.2pt}
\renewcommand{\arraystretch}{1.10}

\resizebox{\textwidth}{!}{
\begin{tabular}{llrrrrrrrrr}
\toprule
\textbf{Object} &
\textbf{Radius [cm]} &
\textbf{Height [cm]} &
\textbf{Standoff [cm]} &
\textbf{Rows/Arm} &
\textbf{Duration [s]} &
\textbf{Rate [Hz]} &
\textbf{Depth$_{110}$} &
\textbf{Depth$_{111}$} &
\textbf{Size [GB]} \\
\midrule

Red Bull
& -- & -- & --
& 1150 & 118.6 & 9.7
& 3412 & 3436 & 14.85 \\

Dumbbell
& 6.90 & 3.66 & 36.9
& 3297 & 110.6 & 29.8
& 3160 & 3194 & 13.78 \\

Controller
& 5.78 & 2.64 & 35.8
& 3093 & 103.6 & 29.9
& 2950 & 2986 & 12.87 \\

Cube
& 4.31 & 1.60 & 34.3
& 2534 & 88.4 & 28.6
& 2522 & 2535 & 10.96 \\

Ketchup
& 4.41 & 14.93 & 34.4
& 4119 & 135.4 & 30.4
& 3921 & 3944 & 17.05 \\

Picture Frame
& 11.60 & 19.07 & 52.2
& 2923 & 99.4 & 29.4
& 2843 & 2867 & 12.37 \\

Mug
& 7.22 & 6.30 & 37.2
& 2954 & 99.4 & 29.7
& 2830 & 2865 & 12.35 \\

Paperweight
& 4.85 & 7.44 & 34.8
& 3465 & 115.4 & 30.0
& 3315 & 3314 & 14.40 \\

Clock
& 5.83 & 9.95 & 35.8
& 3820 & 126.6 & 30.2
& 3656 & 3677 & 15.90 \\

Cup
& 4.11 & 10.16 & 34.1
& 3151 & 107.6 & 29.3
& 3088 & 3107 & 13.43 \\

\midrule

\textbf{Total / Range}
& \textbf{3.5--11.6}
& \textbf{1.6--19.1}
& \textbf{34.1--52.2}
& \textbf{30,506}
& \textbf{1105.1}
& --
& \textbf{31,697}
& \textbf{31,925}
& \textbf{137.94} \\

\bottomrule
\end{tabular}}
\end{table*}

The dataset can consequently be examined at several levels, individual
RGB-D observations, single-arm multi-view sequences, complementary
partial point clouds, or dual-arm fused representations. The raw
recordings remain the primary released data, while the reconstructed
point clouds and meshes presented below are derived products used to
demonstrate geometric consistency.

\subsection{Cross-Arm Geometric Consistency}

A central requirement of the acquisition framework is that independently recorded observations from arm110 and arm111 can be transformed into the same metric coordinate system without requiring visual re-registration of every frame. For each selected observation, the camera pose is obtained from the calibrated base and hand-eye transforms together with the synchronized robot configuration, as described in Eq.~(\ref{eq:world_camera_pose}).

The two arms naturally produce different partial surface measurements because each sensor approaches the object from a different side. In regions observed by both cameras, however, corresponding measurements should occupy similar locations after transformation into the world frame. We therefore use overlap between the independently reconstructed arm-specific point clouds as an additional check of calibration consistency.

Fig.~\ref{fig:cross_arm_alignment} further illustrates the geometric
relationship between the two arm-specific observations. In the shared
world frame, the partial point clouds occupy complementary regions of
the object surface, while the co-observed portions remain reasonably
aligned after calibration and pose recovery.

\begin{figure}[t]
    \centering
    \includegraphics[width=\columnwidth]{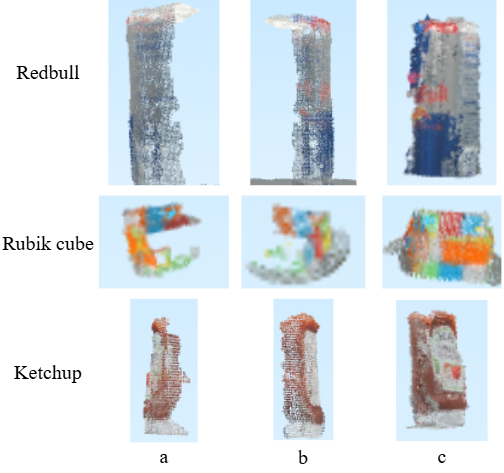}
    \caption{Complementary geometric observations from the two robotic
arms: (a) partial point cloud from arm110, (b) partial point cloud
from arm111, and (c) fused dual-arm point cloud in the shared world
frame.}
    \label{fig:partial_fused}
\end{figure}

The experiments show that the two partial observations align sufficiently well to form coherent combined object surfaces while still preserving complementary geometry. Importantly, the final production fusion does not apply unrestricted ICP or centroid alignment to force the two arms into agreement. The reported fused surfaces are generated primarily from the calibrated camera poses and repeated multi-frame support. As a result, visible agreement between the two arm-specific measurements reflects the quality of the robot-camera calibration and temporal pose association rather than an unconstrained post-hoc registration step.
Fig.~\ref{fig:partial_fused} illustrates the complementary nature
of the two robotic viewpoints. The arm-specific point clouds contain
different visible regions of the object, while their calibrated fusion
combines these measurements into a more complete observed-surface
representation in the shared world frame.

\subsection{Held-Out Surface Evaluation}

To evaluate geometric repeatability without comparing the reconstruction against the same observations used to generate it, a subset of RGB-D keyframes is excluded from fusion. These held-out observations are independently back-projected into 3D and transformed into the world frame using their corresponding robot poses. The resulting query points are then compared with the measured surface mesh using point-to-mesh distance.

Across 224 held-out RGB-D keyframes, the evaluation contains 1,563,466 three-dimensional query points. The aggregate results are summarized in Table~\ref{tab:heldout_results}. The median point-to-mesh distance is 2.13~mm, while the RMSE is 4.04~mm. The 90th-percentile distance is 6.38~mm. In addition, 83.82\% of the held-out points lie within 5~mm of the reconstructed surface and 96.56\% lie within 10~mm.

The values in Table~\ref{tab:heldout_results} should be interpreted as \emph{held-out observation agreement}. They do not represent absolute error against a complete object model obtained from an independent high-accuracy scanner. This distinction is important because the reconstructed mesh contains only surfaces supported by the recorded observations.

\subsection{Qualitative Reconstruction Results}

The qualitative results show the complementary role of the two robotic viewpoints. A point cloud generated from only one arm typically contains the object surfaces visible from that side of the workspace, while self-occluded or oppositely facing regions remain missing. The second arm contributes measurements from a substantially different viewing direction, increasing visible-surface coverage when the observations are combined.

The final fused representation retains repeatedly supported RGB-D measurements and removes isolated observations before surface generation. Multi-scale ball-pivoting produces a mesh over these measured regions without deliberately closing surfaces that were not observed. Together, the held-out evaluation and qualitative reconstructions show that the released calibration, robot-state information, and multimodal measurements are sufficiently consistent to support metric dual-arm fusion.

\begin{table}[t]
\centering
\caption{Held-Out RGB-D Geometric Evaluation}
\label{tab:heldout_results}

\renewcommand{\arraystretch}{1.12}
\setlength{\tabcolsep}{3.5pt}

\begin{tabularx}{\linewidth}{
    >{\raggedright\arraybackslash}p{0.48\linewidth}
    >{\centering\arraybackslash}X
}
\toprule
\textbf{Metric} & \textbf{Result} \\
\midrule

\rowcolor{RowBlue}
Median distance
& \textbf{2.13 mm }\\

\rowcolor{RowGreen}
RMSE
& 4.04 mm\\

\rowcolor{RowOrange}
90th-percentile distance
& 6.38 mm\\

\rowcolor{RowPurple}
Points within 10 mm
& \textbf{96.56\%} \\

\bottomrule
\end{tabularx}
\end{table}

\section{Applications and Comparison with Existing Datasets}

\subsection{Supported Applications}

The dataset is designed as a reusable robotic perception resource rather than for a single reconstruction task. Because RGB, depth, stereo infrared, robot joint states, and calibration information are preserved together, the same acquisition can be represented at the image, point-cloud, trajectory, or fused-surface level.

A direct application is \emph{multi-view 3D reconstruction}. Individual observations from either manipulator provide only partial object geometry, while calibrated observations from both arms can be transformed into the shared world frame to obtain a more complete measured surface. The arm-specific observations also make it possible to study single-arm versus dual-arm perception and quantify the benefit of complementary viewpoints.

The dataset can further support \emph{multimodal robotic perception}. RGB, metric depth, and stereo infrared measurements may be used independently or jointly for object segmentation, recognition, feature fusion, and sensor-ablation studies. Since the corresponding robot configuration and camera pose are retained, perception models can additionally incorporate viewpoint or pose information rather than treating each image as an isolated observation.

Another relevant direction is \emph{active perception and next-best-view planning}. The recorded camera trajectories and partial observations provide examples of how object visibility changes as an eye-in-hand camera moves around the workspace. Future methods could use this information to predict viewpoints that reduce occlusion or provide additional geometric coverage.

The complementary partial observations may also support future learning-based shape completion. Existing methods such as PCN, PoinTr, and related point-cloud completion approaches predict geometry that is absent from the input observation \cite{yuan2018pcn,yu2021pointr}. In contrast, the geometric pipeline in this paper reconstructs only surfaces supported by measured RGB-D observations. Shape completion should therefore be treated as a downstream learning task requiring an appropriate complete-shape target or self-supervised training objective rather than as an output already provided by the dataset.

\begin{table*}[t]
\centering
\caption{Position of the proposed dataset relative to representative RGB-D and robotic object datasets. The comparison emphasizes acquisition characteristics relevant to this work.}
\label{tab:dataset_comparison}

\setlength{\tabcolsep}{6pt}
\renewcommand{\arraystretch}{1.15}

\resizebox{\textwidth}{!}{
\begin{tabular}{lccccc}
\toprule
\textbf{Dataset} &
\textbf{Physical Objects} &
\textbf{RGB-D / Multi-View} &
\textbf{Robotics-Oriented} &
\textbf{Dual Eye-in-Hand Views} &
\textbf{Synchronized Robot State} \\
\midrule

\datasetcell{Washington RGB-D \cite{lai2011rgbd}}
& \yescell{Yes}
& \yescell{Yes}
& \nocell{No}
& \nocell{No}
& \nocell{No} \\

\datasetcell{BigBIRD \cite{singh2014bigbird}}
& \yescell{Yes}
& \yescell{Yes}
& \nocell{No}
& \nocell{No}
& \nocell{No} \\

\datasetcell{YCB \cite{calli2015ycb}}
& \yescell{Yes}
& \yescell{Yes}
& \yescell{Yes}
& \nocell{No}
& \nocell{No} \\

\datasetcell{GraspNet-1Billion \cite{fang2020graspnet}}
& \yescell{Yes}
& \yescell{Yes}
& \yescell{Yes}
& \nocell{No}
& \nocell{No} \\

\datasetcell{ROBI \cite{yang2021robi}}
& \yescell{Yes}
& \yescell{Yes}
& \yescell{Yes}
& \nocell{No}
& \nocell{No} \\

\midrule

\proposedname{DARP (Ours)}
& \proposedyes{Yes}
& \proposedyes{Yes}
& \proposedyes{Yes}
& \proposedyes{Yes}
& \proposedyes{Yes} \\

\bottomrule
\end{tabular}}
\end{table*}

From a robotics perspective, the shared metric representation can also support collaborative manipulation studies. For example, one manipulator may observe surfaces that are occluded from the other, while both observations remain related through a common world frame. Such data can be useful for future systems in which an embodied robot reasons jointly about viewpoint, object geometry, and manipulation actions.

\subsection{Comparison with Existing Datasets}

Table~\ref{tab:dataset_comparison} summarizes the position of the proposed dataset relative to representative object-centric RGB-D and robotic perception resources discussed in Section~II. The comparison is intended to highlight differences in acquisition design rather than to suggest that the datasets address identical research problems.

The Washington RGB-D Object Dataset and BigBIRD provide multi-view RGB-D observations of physical objects for object recognition and 3D perception \cite{lai2011rgbd,singh2014bigbird}. The YCB Object and Model Set combines physical manipulation objects with associated visual and geometric models and has become widely used in robotics research \cite{calli2015ycb}. GraspNet-1Billion connects RGB-D observations with large-scale grasp annotations \cite{fang2020graspnet}, while ROBI focuses on multi-view observations of reflective objects in robotic bin-picking environments \cite{yang2021robi}.

The proposed dataset differs primarily in the way observations are acquired and geometrically linked. Two independently moving eye-in-hand cameras observe the same object from opposite sides of a shared manipulation workspace, while synchronized robot joint states and calibration parameters are retained with the original sensor recordings. This allows the pose of each selected observation to be reconstructed through robot kinematics and enables direct analysis of complementary arm-specific and dual-arm measurements.

The purpose of this comparison is not to claim that existing datasets lack multi-view or robotic information. Instead, the distinction of the proposed dataset is the combination of two independently moving eye-in-hand sensors, continuous multimodal acquisition, synchronized manipulator states, and a calibration chain that expresses both camera streams in the same metric world frame. This combination supports direct study of complementary robotic viewpoints while preserving the raw acquisition data for alternative processing and learning methods.

\section{Discussion and Limitations}

The dataset demonstrates that complementary observations from two independently moving eye-in-hand cameras can be aligned in a shared metric frame using robot calibration, synchronized joint states, and forward kinematics. However, several limitations should be considered.

First, the reconstructed meshes represent only \emph{observed surface geometry}. Regions that are not visible to either camera, such as undersides or strongly occluded surfaces, may remain incomplete. The current geometric pipeline does not infer or artificially close these missing regions. Second, reconstruction quality depends on the calibration state of the robotic system. Changes in camera mounting, robot-base position, sensor brackets, or joint-zero behavior may reduce cross-arm consistency and require recalibration. Third, the current release contains ten unique tabletop objects collected in a single laboratory setup. Although the objects provide geometric variation, the dataset does not cover the full diversity of materials, shapes, and environmental conditions encountered in broader robotic applications.

Finally, the held-out point-to-mesh evaluation measures agreement with unused RGB-D observations from the same acquisition rather than absolute error against an independently scanned complete-object model. Future extensions may include additional objects, broader sensing conditions, reference 3D models, and learning-based studies such as shape completion and next-best-view selection.

\section{Conclusion}

This paper presented a calibrated dual-arm RGB-D-IR dataset for object-centered robotic perception using two independently moving eye-in-hand manipulators. The dataset preserves continuous multimodal sensor recordings, synchronized robot joint states, object-level metadata, and calibration information so that observations from both arms can be reconstructed in a shared metric frame. To demonstrate the geometric consistency of the data, we used a deterministic multi-view fusion pipeline that transforms calibrated RGB-D measurements into complementary partial point clouds and measured surface meshes without using learned completion or generative methods. Evaluation with held-out RGB-D observations produced a median point-to-mesh distance of 2.13~mm, with 96.56\% query points within 10~mm of the reconstructed surface. The dataset currently contains ten unique tabletop objects and is intended as a reusable acquisition resource rather than a task-specific benchmark. Future work will extend the object and sensing diversity and investigate downstream applications including multimodal perception, next-best-view planning, collaborative manipulation, and learning-based reasoning over partially observed object geometry.

\section*{Acknowledgment}

The authors acknowledge the support and resources provided by the
Bioinspired Robotics, AI, Imaging and Neurocognitive Systems (BRAINS)
Laboratory at The University of Alabama.

\bibliographystyle{IEEEtran}
\bibliography{paper_references}

\end{document}